\documentclass{article} 
\usepackage{iclr2027_conference,times}

\usepackage{amsmath,amsfonts,bm}

\def\eqref#1{equation~\ref{#1}}

\def\1{\bm{1}}

\DeclareMathAlphabet{\mathsfit}{\encodingdefault}{\sfdefault}{m}{sl}
\SetMathAlphabet{\mathsfit}{bold}{\encodingdefault}{\sfdefault}{bx}{n}

\usepackage{url}
\usepackage{url}
\usepackage{booktabs}
\usepackage{graphicx}
\usepackage{tabularx}
\usepackage{longtable}
\usepackage{xcolor}
\usepackage{colortbl}
\usepackage{graphicx}
\usepackage{wrapfig}
\usepackage{xspace}
\usepackage[T1]{fontenc}
\usepackage{tcolorbox}
\usepackage{hyperref}
\hypersetup{
  colorlinks=false,
  pdfborder={0 0 1},
  citebordercolor={0 1 0}
}
\tcbuselibrary{skins,breakable,listings}

\definecolor{PromptBlue}{HTML}{356A96}
\definecolor{PromptTeal}{HTML}{287D78}
\definecolor{PromptPurple}{HTML}{76549B}
\definecolor{PromptAmber}{HTML}{A66B2B}

\newtcblisting{promptbox}[2]{
  enhanced,
  breakable,
  listing only,
  colback=#1!4!white,
  colframe=#1,
  colbacktitle=#1!12!white,
  coltitle=#1!35!black,
  title={#2},
  fonttitle=\small\bfseries,
  boxrule=0.6pt,
  arc=2mm,
  left=2.5mm, right=2.5mm,
  top=2mm, bottom=2mm,
  before skip=10pt, after skip=10pt,
  listing options={
    basicstyle=\ttfamily\footnotesize,
    columns=fullflexible,
    breaklines=true,
    breakatwhitespace=false,
    showstringspaces=false,
    keepspaces=true,
    tabsize=2
  }
}

\title{Where Do Test-Time Scaling and Training Fall Short in Individual Stance Prediction?}

\author{
Yuyang Zhao\thanks{Equal contribution.} \quad
Xuan Liu$^{1}$\footnotemark[1]\ \ \thanks{Corresponding.} \quad
HaoYang Shang\footnotemark[1] \quad
Haojian Jin$^{1}$ \\
$^{1}$University of California, San Diego
}

\newcommand{\benchmark}{\textsc{Stance-Bench}\xspace}

\usepackage{pifont}
\usepackage{tikz}
\usepackage{xcolor}

\DeclareRobustCommand{\cmark}{\textcolor{green!55!black}{\ding{51}}}
\DeclareRobustCommand{\xmark}{\textcolor{red!70!black}{\ding{55}}}
\DeclareRobustCommand{\pmark}{\textcolor{orange!85!black}{%
  \tikz[baseline=-0.6ex]{%
    \draw (0,0) circle (0.8ex);%
    \fill (0,0.8ex) arc (90:270:0.8ex) -- cycle;%
  }}%
}

\begin{document}

\iclrfinalcopy

\maketitle
\lhead{Preprint.} 

\begin{abstract}
Test-time scaling and post-training have improved LLM performance in coding and mathematical reasoning, but their effectiveness for individual stance prediction remains unclear. We study this question by predicting a person's stance in a new discussion from their history. We evaluate widely used test-time scaling strategies and post-training methods, such as supervised fine-tuning and reinforcement learning, and identify four failure modes across generation, selection, and learning: (1) \emph{incorrect consensus}, where repeated samples agree on the wrong stance; (2) \emph{selection failure}, where generation covers the observed stance but selection misses it; (3) \emph{response overfitting}, where supervised fine-tuning improves imitation but harms prediction; and (4) \emph{early plateau}, where reinforcement learning shows modest initial gains followed by limited further improvement. We expose these failures using \benchmark, which contains 2499 prediction tasks from 500 Hacker News users. Guided by this analysis, we explore a simple approach that combines direct scores for all candidate stances with explicit assessments of support from the individual's history. On the 781-task test set, this approach achieves 21.83 discussion-specific Macro F1 with Qwen3-8B, compared with 19.27 for direct scoring.
Our results motivate evaluating candidate generation, final selection, and person-specific evidence use separately.  Our data is available at \href{https://github.com/stance-bench/Stance-Bench}{https://github.com/stance-bench/Stance-Bench}.
\end{abstract}

\section{Introduction}

Large language models (LLMs) are increasingly used to simulate human opinions and decisions \citep{argyle2023out,chuang2024beyond,liu2025exploring}, as well as interactions among social agents \citep{park2023generative,zhou2024sotopia}, motivating their broader use in social simulation \citep{gao2024large,anthis2025position}. Their appeal is clear: simulated participants could make behavioral research faster, cheaper, and easier to scale, while behavioral predictions could help anticipate how populations and individuals respond to new issues and interventions. Their fidelity, however, remains limited at both population and individual levels \citep{hu2026simbench,dominguez2024questioning,bisbee2024synthetic,liu2026humanstudy}. At the individual level, predictions of a particular person's held-out responses and behavioral trajectories remain unreliable, even when models are conditioned on demographic information or prior behavior \citep{chen2026synthetic,lu2026can,liu2026cobra}. These limitations contrast with the rapid gains from two widely used approaches to improving LLM capabilities: test-time scaling (TTS) for reasoning \citep{wang2023selfconsistency,snell2025scaling}, and reinforcement learning (RL) post-training for mathematical reasoning and code generation \citep{deepseekai2025deepseekr1}.

This paper examines whether gains from TTS and post-training extend to individual stance prediction and where their limitations arise. TTS uses additional inference-time computation to generate, aggregate, or select among candidate answers, while post-training adapts model behavior using demonstrated responses or reward feedback \citep{snell2025scaling,wang2023selfconsistency,ouyang2022training}. Individual stance prediction is important for public-opinion polling and survey research \citep{argyle2023out,park2024llm}, as well as market research and policy decision-making \citep{chen2026synthetic}. In this setting, the most common or plausible stance in a discussion may not be the stance of the particular person being predicted. This raises the question of whether methods that improve answer generation and selection also improve individual prediction. Answering this question requires more than reporting end-to-end accuracy. For inference-time methods, we distinguish failure to generate the target stance from failure to select it once generated. For training-based methods, we examine whether response imitation and outcome rewards provide useful signals for person-specific prediction.

\begin{figure}[t]
    \centering
    \includegraphics[width=\textwidth]{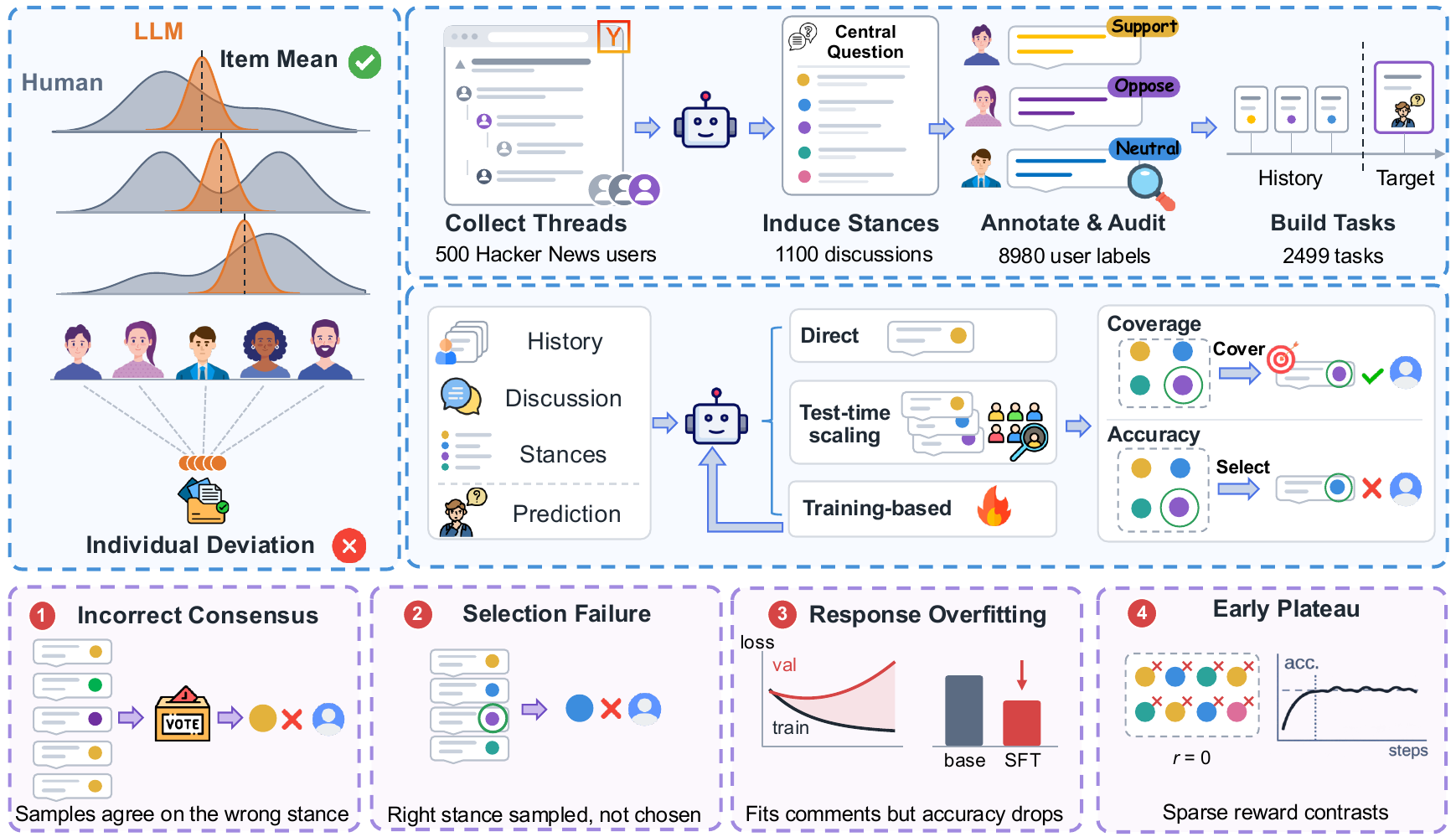}
    \caption{\textbf{Overview.} \benchmark is built by collecting Hacker News threads, inducing discussion-specific stances, and annotating user contributions. Given a user's history, the discussion, and its stance options, a model predicts the user's stance. We identify four failure modes: \emph{incorrect consensus} (samples agree on a wrong stance), \emph{selection failure} (the observed stance is generated but not selected), \emph{response overfitting} (SFT fits responses better but predicts stances worse), and \emph{early plateau} (RL gains little after early training, and reward contrast stays sparse).}
    \label{fig:overview}
    \vspace{-3mm}
\end{figure}

To examine these methods at multiple stages, we construct \benchmark as a diagnostic instrument. It contains $2{,}499$ prediction tasks from $500$ Hacker News users across $1{,}100$ annotated discussions. Given a target discussion and a user's preceding history, a model predicts the contribution that user would make, conditional on participation. Discussion-specific annotations map open-ended generated comments and observed contributions to a shared stance space. This design lets us evaluate final predictions while separately measuring whether the target stance appeared among intermediate candidates.

Our analysis identifies four failure modes across generation, selection, and learning. First, \textbf{incorrect consensus}: repeated sampling often converges on the wrong stance, leaving the target stance absent from the candidate pool. Second, \textbf{selection failure}: broader exploration improves coverage, but the observed stance is frequently not selected even when generated. Third, \textbf{response overfitting}: response-level supervised fine-tuning improves imitation of training sequences while degrading stance prediction. Fourth, \textbf{early plateau}: exact-match RL yields modest initial gains with limited subsequent improvement. Most rollout groups provide no reward contrast, and the share with nonzero reward advantages declines between the first and final training windows. These findings show how additional computation, response imitation, and outcome rewards can fail at different stages of individual prediction.

Guided by this analysis, we explore an explicit evidence reader that evaluates a user's historical comments against competing stances while keeping the language model frozen. Its Macro F1 improvement over direct option scoring, together with history-replacement controls, provides initial evidence that explicitly evaluating personal evidence can help. We treat this method as a direction suggested by the failure analysis rather than a complete solution to the four failure modes.

Our contributions are threefold:
\begin{itemize}
    \item \textbf{Failure-mode analysis.} We analyze widely used test-time scaling and post-training methods for individual stance prediction, separating errors in candidate generation, selection, response imitation, and reward-based learning.
    \item \textbf{Diagnostic evaluation.} We introduce \benchmark, which connects open-ended generation to observed individual stances and records intermediate candidates and final predictions separately.
    \item \textbf{Analysis-guided method.} We explore an evidence-reading method motivated by the analysis, with controls testing whether its predictions depend on evidence from the target individual.
\end{itemize}









\section{Motivation}
\label{sec:motivation}

\subsection{From General Improvements to Individual Prediction}
\label{sec:motivation-transfer}

Test-time scaling and post-training have improved performance on many
reasoning tasks
\citep{wang2023selfconsistency,snell2025scaling,shao2024deepseekmath}.
Whether these gains extend to individual prediction is an empirical
question. In individual stance
prediction, multiple responses may be plausible, while evaluation asks
which stance matches the target individual. Overall response quality and
person-specific predictive accuracy are therefore distinct quantities.
\noindent\textbf{Test-time scaling.}
Test-time methods generate additional candidates and then aggregate,
rank, or select among them
\citep{wang2023selfconsistency,chen2023universal,ni2023lever}. Their gains
require both generating the target stance and selecting it: agreement
does not by itself imply accuracy, and diverse candidates still require
a selection rule.
\noindent\textbf{Supervised fine-tuning.}
Supervised fine-tuning increases a model's fit to demonstrated responses
\citep{ouyang2022training}. Here, the training target contains a comment
and its stance label, while evaluation uses the label. Whether improved
response fit also improves held-out stance prediction must be evaluated
directly.
\noindent\textbf{Reinforcement learning.}
Outcome-based reinforcement learning encourages responses that receive
higher task rewards \citep{shao2024deepseekmath}. Group-relative updates
use reward differences among sampled outputs, so learning from stance
correctness depends in part on contrast within a group. We therefore track
both predictive performance and the fraction of rollout groups that
provide nonzero reward advantages.

These mechanisms motivate stage-specific evaluation without assuming
in advance how the methods will behave.

\subsection{What the Evaluation Must Reveal}
\label{sec:motivation-diagnosis}

End-to-end accuracy alone cannot characterize these stages. Higher accuracy
may reflect better recognition of common positions, better use of the
target person's history, or both. Conversely, a method may respond to
personal evidence without producing a large aggregate improvement. An
evaluation must therefore reveal both where a prediction
fails and whether it depends on information from the correct person.
\noindent\textbf{Shared prediction context (R1).}
Multiple users should be observed in the same discussion. Holding the
discussion fixed makes their positions directly comparable and helps
distinguish a generally likely response from a response associated
with a particular person.
\noindent\textbf{Controlled personal evidence (R2).}
The target person's history should be removable or replaceable while
the discussion and prediction target remain unchanged. Comparing the
correct history with another person's history tests
whether performance depends on evidence matched to the individual,
rather than on the presence of additional text.
\noindent\textbf{Candidate and selection traces (R3).}
Intermediate candidates and final predictions should be recorded
separately within a shared stance space. This makes it possible to
determine whether a change in final performance arises during candidate
generation or during the final decision.

Together, these requirements connect changes in final performance to
candidate generation, final selection, and the use of person-specific
evidence.

\subsection{What Existing Evaluations Establish}
\label{sec:motivation-paradigms}

\noindent\textbf{Population-level simulation.}
OpinionQA and SimBench evaluate population-level correspondence and
simulation fidelity \citep{santurkar2023whose,hu2026simbench}.
Distributional agreement, however, does not establish that positions
are assigned to the correct individuals.
\noindent\textbf{Individual simulation and personalization.}
Interview-, survey-, and history-conditioned evaluations establish the
value of personal information~\citep{park2024llm,salemi2024lamp}, but
do not always isolate whether performance depends on matching evidence
to the correct individual.
\noindent\textbf{Stance detection and prediction.}
SemEval-2016 Task~6 evaluates stance identification from observed text
\citep{mohammad2016semeval}, while \citet{loh2024predicting} predict
stances from target-agnostic user posts. Neither separately exposes candidate
generation and final selection within naturally occurring shared
discussions.

Existing evaluations therefore illuminate important parts of the
problem, but do not expose all of these stages within one setting.
Section~\ref{sec:benchmark} describes the
evaluation framework we use to connect shared contexts, controlled
personal evidence, intermediate candidates, and final predictions. Appendix~\ref{app:diagnostic-coverage} provides a comparison of existing evaluations and \benchmark against these diagnostic requirements.

\section{Benchmark Construction}
\label{sec:benchmark}

\begin{wrapfigure}{R}{0.4\textwidth}
\centering
\vspace{-\intextsep}
\includegraphics[width=0.9\linewidth]{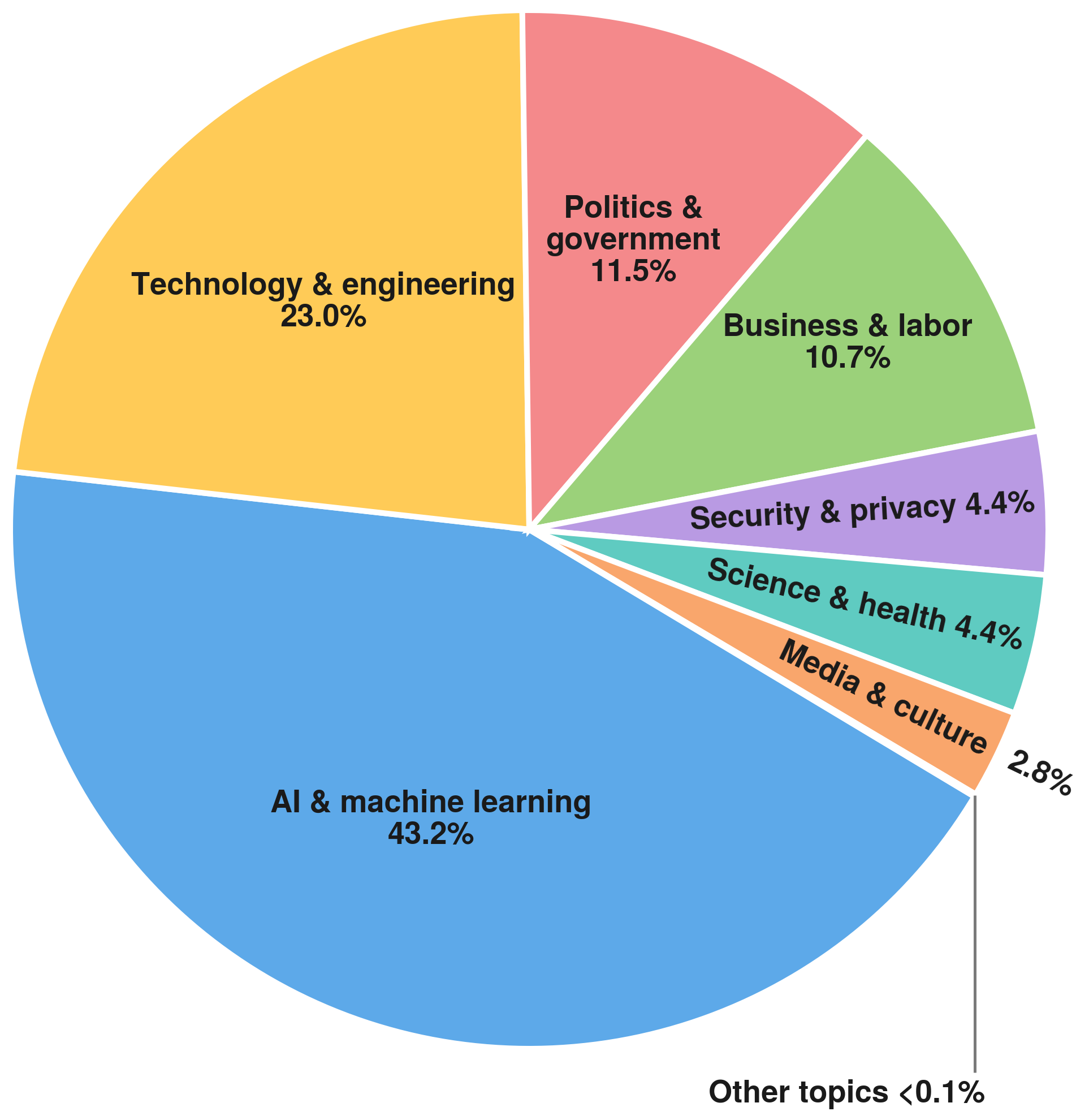}
\small
\setlength{\abovecaptionskip}{5pt}
\setlength{\belowcaptionskip}{0pt}
\caption{Topic composition of all 2499 \benchmark tasks.}
\label{fig:benchmark-composition}
\end{wrapfigure}

\textbf{Task Design.}\label{sec:task-design} Given a user's preceding history $H_u$, a target discussion $X_d$, and candidate stances $\mathcal{L}_d$, models predict the user's dominant stance conditional on participation. Models generate a natural-language comment and a stance label, and the evaluation uses the label. Candidate stances answer the discussion's central question, with explicit definitions, exclusion boundaries, and supporting paraphrases.

We evaluate exact-match accuracy and discussion-specific Macro F1 on targets assigned a listed stance. Macro F1 averages equally over reference-supported stances, treating stances from different discussions as distinct classes and using a common class set across methods. For multiple-candidate methods, coverage is the fraction of tasks whose observed stance appears among candidates; label diversity is the mean number of distinct candidate stances.


\noindent\textbf{Data Collection.}\label{sec:data-collection} We collect public histories for 500 active Hacker News users through the Algolia index and official API. We select shared discussions to capture histories across discussions and differences between users responding to the same context.

GPT-5.5 \citep{openai2026gpt55} identifies each discussion's central question and candidate stances, initially excluding comments from designated target users. Target contributions are later used to assign reference labels and to audit whether the shared taxonomy covers observed positions, but never appear in model-visible inputs. Revisions from these audits are applied at the discussion level rather than tailored to a single target. We retain the supporting evidence for each annotation.

Tasks use up to five recent discussions per user. Historical discussions qualify only if the user's final comment precedes their first comment in the target discussion. We remove cohort-authored comments from model-visible target context, pseudonymize users, and store prediction inputs separately from reference outcomes.

\noindent\textbf{Dataset Statistics and Quality.}\label{sec:dataset-statistics} \benchmark comprises 1100 annotated discussions and 8980 user-discussion annotations. Its 2499 prediction tasks span 414 target discussions, with four or five tasks per user. User histories cover 9 to 17 discussions with the median of 15. Taxonomies contain 4 to 10 candidate stances with the mean of 5.94.

\section{Failure Modes and Limits of Transfer}
\label{sec:failure-modes}

We identify four failure modes in the evaluated inference-time and post-training methods. \textbf{Incorrect consensus} occurs when repeated predictions agree on a stance other than the target user's. \textbf{Selection failure} occurs when the observed stance appears among candidates but is not selected. \textbf{Response overfitting} occurs when supervised training fits response sequences better without improving stance prediction. Continued training then overfits the training responses. \textbf{Early plateau} describes modest initial gains from reward-based optimization followed by limited additional improvement. These findings expose distinct limitations in generating, selecting, and learning to attribute a stance to an individual. Method settings are given in Appendices~\ref{app:methods} and~\ref{app:training}.

\subsection{Incorrect Consensus}
\label{sec:incorrect-consensus}

\begin{figure*}[t]
    \centering
    \begin{minipage}[t]{0.32\textwidth}
        \centering
        \includegraphics[width=\linewidth]{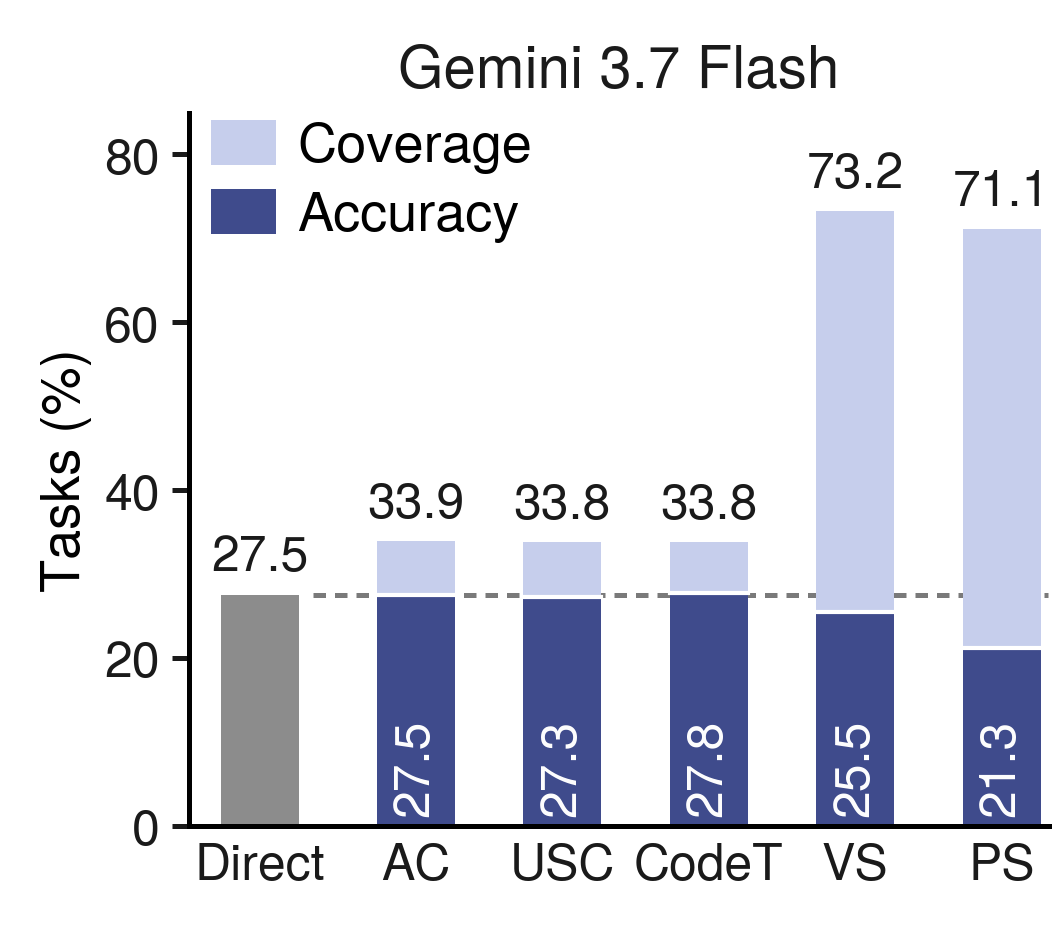}
        \par\vspace{2pt}
        {\small (a)}
    \end{minipage}\hfill
    \begin{minipage}[t]{0.32\textwidth}
        \centering
        \includegraphics[width=\linewidth]{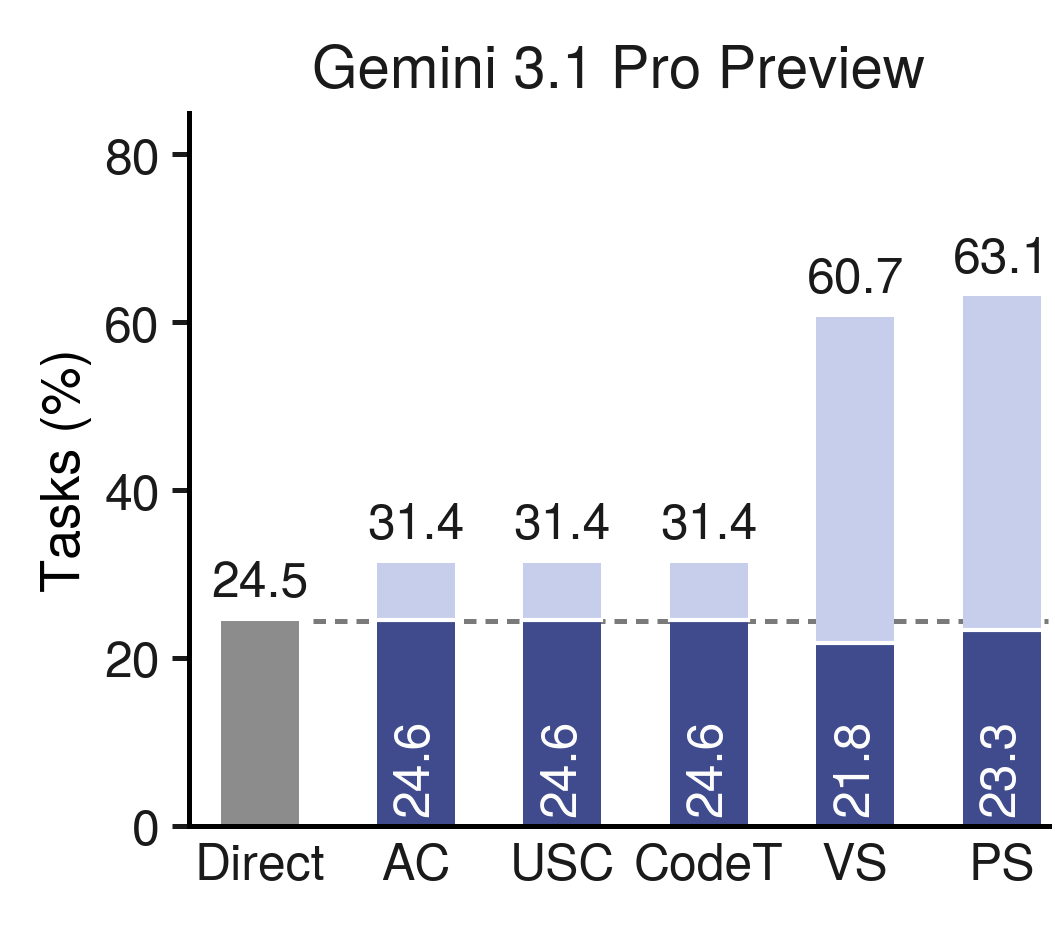}
        \par\vspace{2pt}
        {\small (b)}
    \end{minipage}\hfill
    \begin{minipage}[t]{0.32\textwidth}
        \centering
        \includegraphics[width=\linewidth]{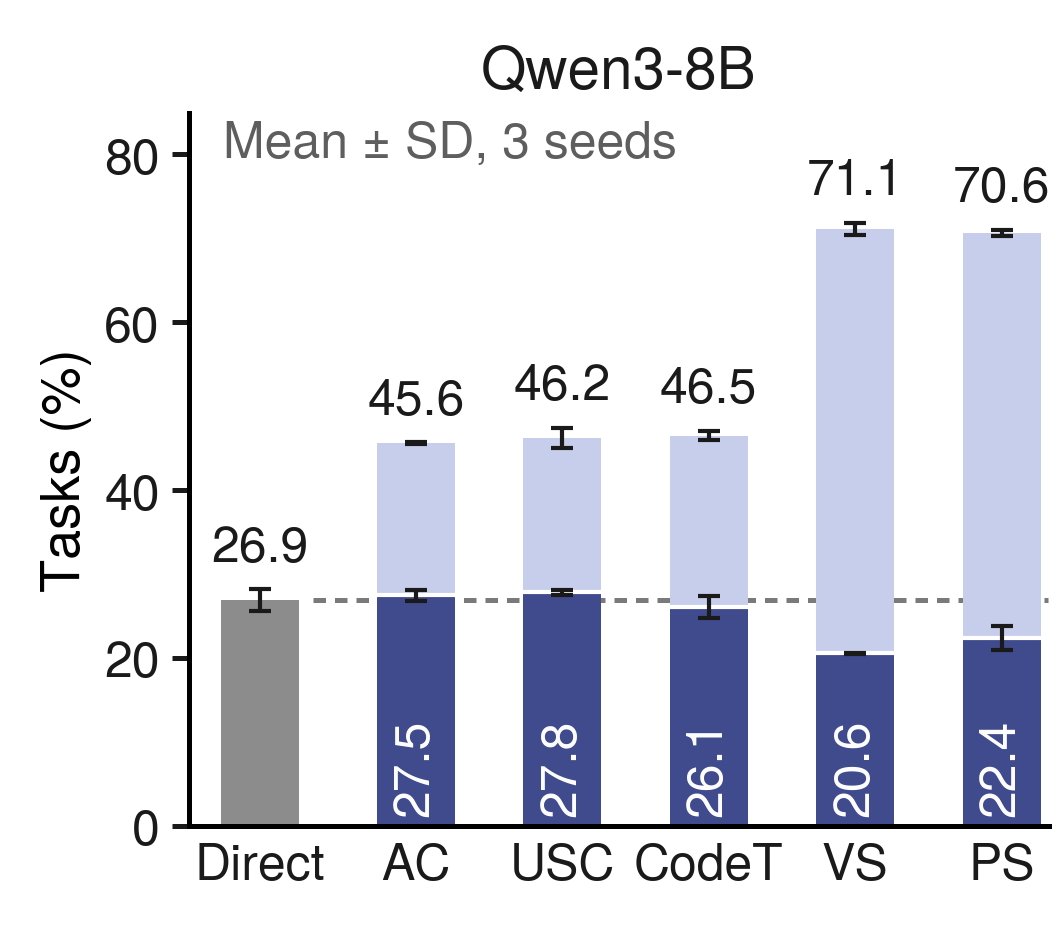}
        \par\vspace{2pt}
        {\small (c)}
    \end{minipage}
    \caption{Candidate coverage and final prediction accuracy for
    (a) Gemini 3.7 Flash, (b) Gemini 3.1 Pro, and (c) Qwen3-8B. Coverage measures whether the target user's observed stance appears among the generated candidates and accuracy measures whether the final prediction matches that stance. Their difference identifies cases in which the observed stance is available but not selected.}
    \label{fig:failure-candidates-revised}
    \vspace{-3mm}
\end{figure*}

Repeated agreement was often agreement on the wrong stance. In our AlphaCode-style consensus adaptation \citep{li2022alphacode}, Gemini Flash produced only 1.29 distinct stance labels across four candidates, and candidate coverage was 33.9\%. Universal Self-Consistency (USC; \citealp{chen2023universal}) showed nearly identical diversity and coverage. Thus, on approximately two-thirds of targets, none of the sampled labels matched the user's observed stance shown in Figure~\ref{fig:failure-candidates-revised}(a). Any selector restricted to these candidates was already unable to recover the correct prediction.

Incomplete coverage was also observed for Gemini Pro (31.4\%) and Qwen3-8B (45.6\%; Figure~\ref{fig:failure-candidates-revised}(b,c)). Qwen consensus nevertheless improved on direct generation (Table~\ref{tab:tts-main}).

\subsection{Selection Failure}
\label{sec:selection-failure}

Generating the observed stance did not ensure that it was selected. With Gemini Flash, Verbalized Sampling \citep{zhang2025verbalized} increased coverage to 73.2\%, yet selected the correct stance on only 34.8\% of covered targets, despite producing nearly four distinct labels per task. This rate is only modestly above the one-in-four expected from a uniform choice among four distinct labels, so the model's verbalized probabilities carried little information about which candidate matched the user. The large light-blue segments in Figure~\ref{fig:failure-candidates-revised} separate this selection loss from candidate omission. The same pattern appeared with Gemini Pro (35.9\% of covered targets) and in the three-seed Qwen results (28.9\%).

Higher coverage therefore did not yield higher accuracy. On all three backbones, Verbalized Sampling covered the observed stance on more than twice as many targets as direct generation, yet its accuracy was lower (Table~\ref{tab:tts-main}). Consensus methods showed the opposite trade-off. With Flash, Universal Self-Consistency selected the observed stance on 81\% of covered targets but covered only 33.8\% of targets. Selecting by agreement works among similar candidates but limits coverage (Section~\ref{sec:incorrect-consensus}), whereas diverse candidates raise coverage without a reliable way to choose among them.

\subsection{Response Overfitting}
\label{sec:sft-failure}

\begin{figure*}[h]
    \centering
    \begin{minipage}[t]{0.32\textwidth}
        \centering
        \includegraphics[width=\linewidth]{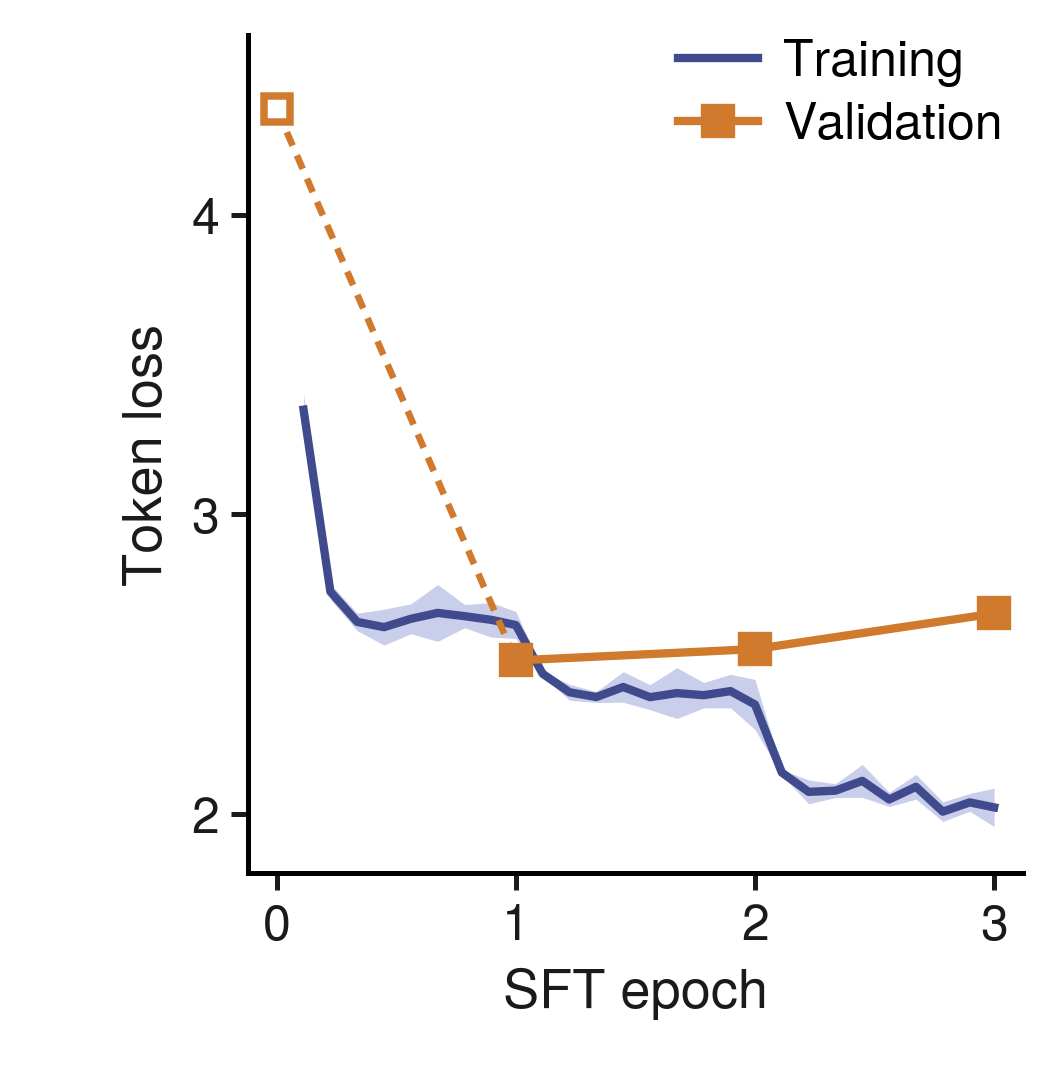}
        \par\vspace{2pt}
        \vspace{-2mm}
        {\small (a)}
    \end{minipage}\hfill
    \begin{minipage}[t]{0.32\textwidth}
        \centering
        \includegraphics[width=\linewidth]{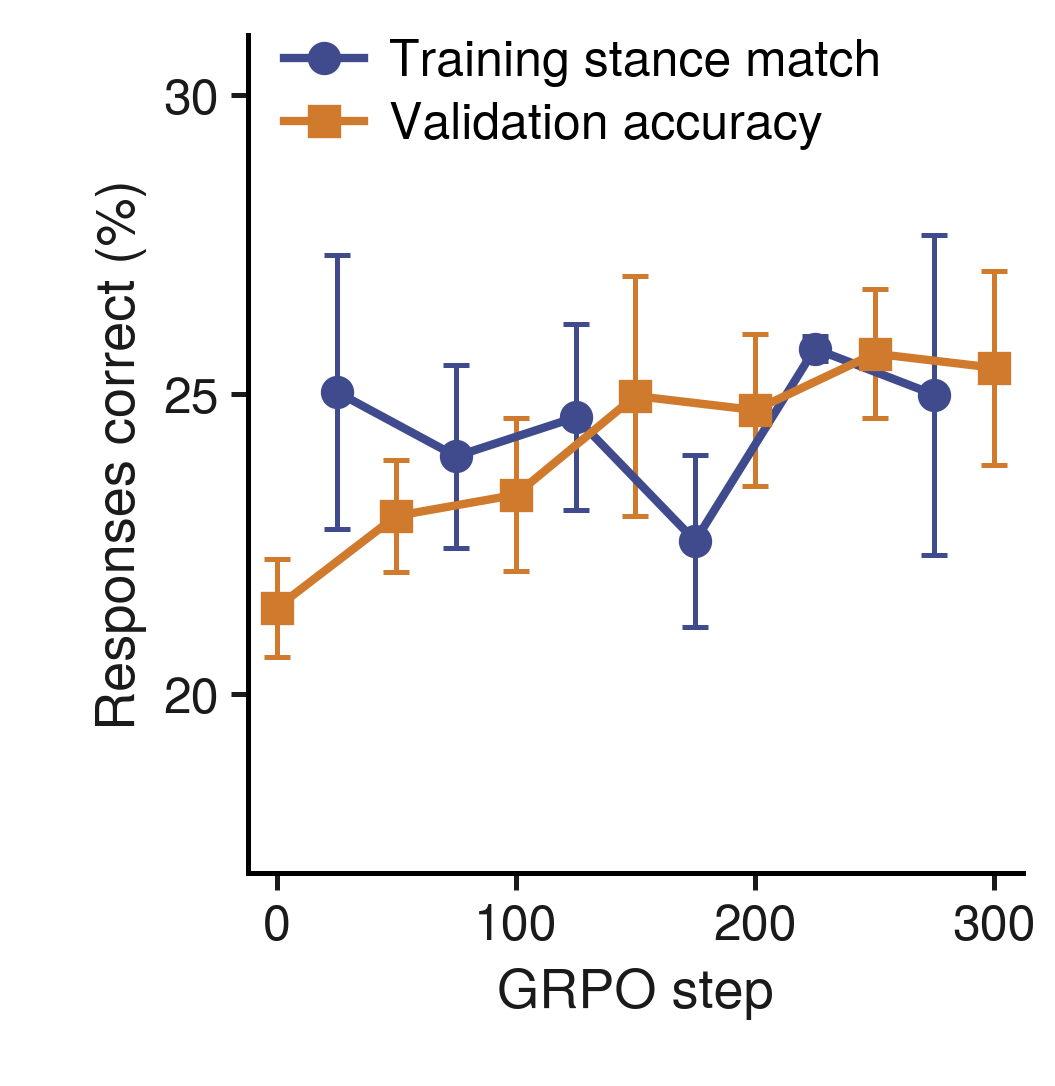}
        \par\vspace{2pt}
        \vspace{-2mm}
        {\small (b)}
    \end{minipage}\hfill
    \begin{minipage}[t]{0.32\textwidth}
        \centering
        \includegraphics[width=\linewidth]{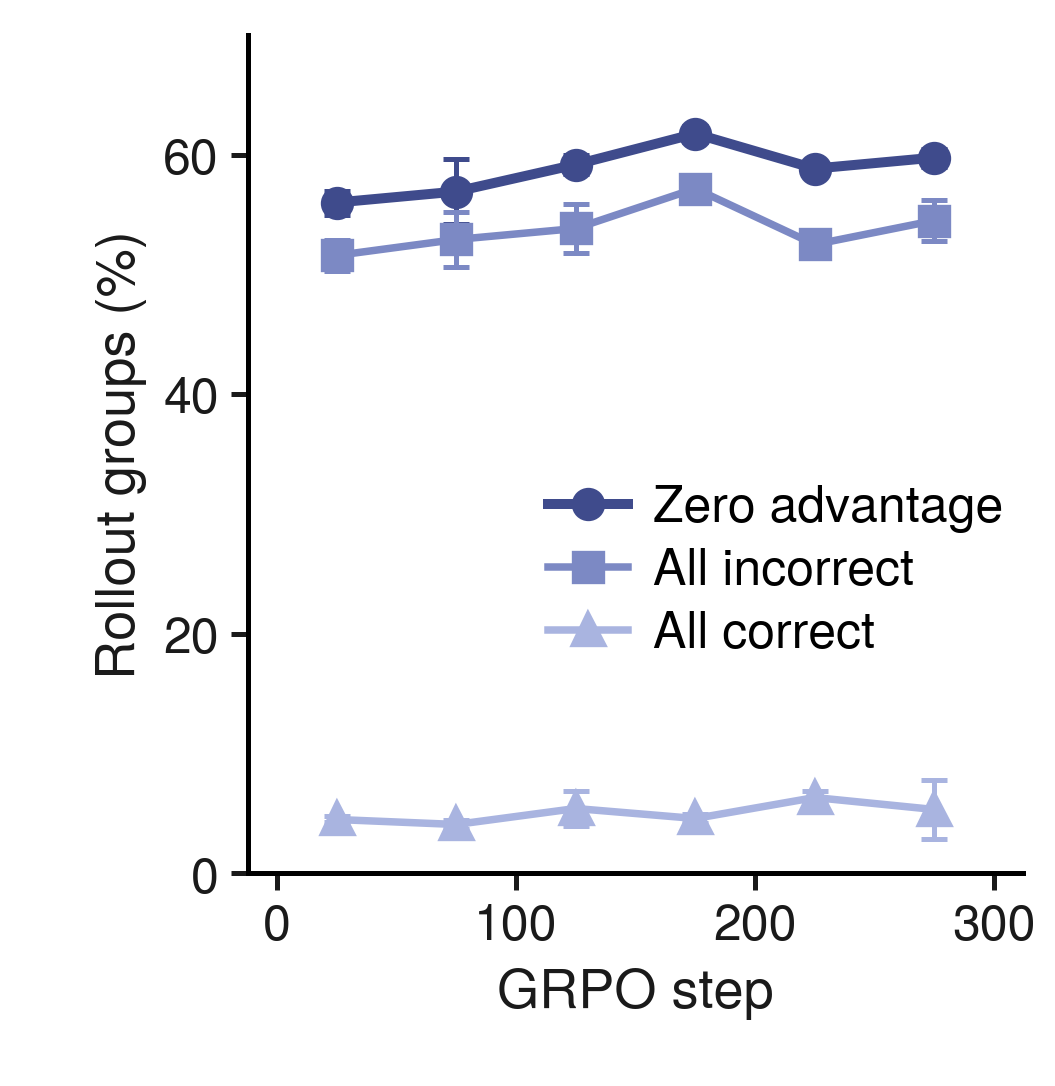}
        \par\vspace{2pt}
        \vspace{-2mm}
        {\small (c)}
    \end{minipage}
    \vspace{-2mm}
    \caption{\textbf{Training diagnostics for Qwen3-8B.} (a) Training and validation average token loss during comment-and-label SFT with 3 random seeds. (b) Sampled training stance-match rate and first-response validation accuracy during GRPO. (c) Fractions of eight-response rollout groups with all responses incorrect, all correct, or zero reward advantages. Curves show means across three training seeds; shading and error bars indicate $\pm 1$ sample standard deviation. The SFT epoch-0 point is a shared base-model evaluation without an error bar. GRPO rollout statistics use non-overlapping 50-step windows.}
    \label{fig:training-diagnostics}
\end{figure*}

Comment-and-label SFT improved response fit but not stance prediction. Mean validation token loss fell from 4.35 at initialization to 2.51 after one epoch (Figure~\ref{fig:training-diagnostics}(a)). We trained with three random seeds and selected one checkpoint per seed by validation stance F1. These checkpoints reached 19.50\% mean test stance accuracy, compared with 27.78\% for the shared decoding-matched base control. Lower token loss on real comments did not translate into more accurate stance assignment. Continued training then overfit the training responses. In all three seeds, training loss kept decreasing after the first epoch while validation loss rose, reaching a mean of 2.67 at epoch three.

SFT also produced more \texttt{OTHER} and invalid outputs on the listed-position targets. \texttt{OTHER} is a valid training label but counts as an error on this subset. Some invalid responses reached the generation limit before producing a label. Fitting response sequences thus co-occurred with incorrect stance assignments and failures to produce usable predictions.

\subsection{Early Plateau}
\label{sec:rl-failure}

Exact-match GRPO with KL regularization produced modest early gains, followed by limited additional improvement (Figure~\ref{fig:training-diagnostics}(b)). Mean first-response accuracy on 283 fixed validation targets rose from 21.4\% at initialization to 25.0\% at step 150. It reached 25.4\% at step 300, adding less than half a percentage point over the second half of training. All three seeds improved from initialization by 1.8, 6.4, and 3.9 percentage points, respectively. Accuracy on 200 fixed training targets also increased by 2.0--3.5 points between steps 0 and 250, the last available training-set evaluation. Across the three 300-step runs, most of the mean validation improvement occurred within the first 150 steps.

Within-group reward contrast remained sparse throughout training (Figure~\ref{fig:training-diagnostics}(c)). The fraction of groups with nonzero reward advantages was lower in the final 50-step window than in the first for every seed. Its mean was 44.0\% in the first window and 40.3\% in the last, with fluctuations in between. Almost all groups without reward contrast were all-wrong or all-correct groups of eight responses; format penalties account for small differences between these statistics. Equal rewards yield zero reward advantages, but the KL term can still update the policy. Sparse reward contrast co-occurred with the early plateau in these runs.

%
\section{Exploring Explicit Personal Evidence}
\label{sec:method}

Section~\ref{sec:failure-modes} shows two problems: the target user's
stance is often absent from generated candidates, and it is often not
selected even when present.
We explore a simple response: score every listed stance directly, read
the user's own history explicitly against those stances, and combine
the two with two fitted coefficients. We present this as a probe guided
by the failure analysis rather than as a complete solution.

\paragraph{Direct scores.}
Incorrect consensus shows that selecting from a sampled pool is bounded
by candidate coverage (Section~\ref{sec:incorrect-consensus}). We
therefore score every listed stance by the frozen model's label
log-likelihood,
\begin{equation}
B_j=\log \pi_\theta(j\mid X_d,H_u),\qquad j\in\mathcal{L}_d .
\label{eq:direct-score}
\end{equation}
This makes every stance eligible for prediction but leaves the ranking
unchanged: $\arg\max_j B_j$ is still the model's own preferred stance.

\paragraph{Evidence scores.}
Selection failure shows that availability does not establish
attribution (Section~\ref{sec:selection-failure}). We therefore add an
evidence score read explicitly from the user's history. For each
historical comment $c_i$, a frozen reader returns a distribution $p_i$
over $\mathcal{L}_d$ and a relevance weight $w_i\in[0,1]$, the
probability that the comment addresses the question at all. Following
logarithmic opinion pooling \citep{genest1984aggregating}, we
aggregate
\begin{equation}
S_j=\sum_i \bar w_i\log p_i(j),\qquad \bar w_i=\frac{w_i}{\sum_h w_h}.
\label{eq:evidence-score}
\end{equation}
Comments judged irrelevant receive little weight, and only differences
$S_j-S_k$ affect the ranking, so a comment that supports two stances
equally leaves their order unchanged. Retrieval and reader details are
given in Appendix~\ref{app:method}.

\paragraph{Fusion.}
We combine the two scores log-linearly,
\begin{equation}
P(y=j\mid u,d)=\operatorname{softmax}_{j\in\mathcal{L}_d}\big(aB_j+bS_j\big),
\label{eq:final-fusion}
\end{equation}
and predict the most probable stance. Setting $b=0$ leaves only the
direct scores. The coefficients $a$ and $b$ set the relative weight of the
model's preference and the user's evidence, and thus determine when
evidence can overturn the model's preferred stance. For fixed $a$ and
$b$, the softmax does not change which stance ranks first; its role is
to turn the combined scores into probabilities, so that $a$ and $b$
can be learned by maximizing the probability assigned to observed
stances on development targets (Appendix~\ref{app:method}).
Supervision therefore acts directly on stance discrimination, rather
than through response imitation or outcome rewards
(Sections~\ref{sec:sft-failure} and~\ref{sec:rl-failure}), while all
language-model parameters remain fixed.
Eq.~\ref{eq:final-fusion} can also be read as a multinomial logit
model of discrete choice \citep{McFadden1974}.

\section{Experiments}
\label{sec:overall-performance}

Following the diagnostics in Section~\ref{sec:failure-modes}, we evaluate whether additional test-time computation and learned adaptations improve individual stance prediction. Our method achieves higher accuracy and Macro F1 than direct generation and direct scoring under the same evaluation protocol.

Our main comparisons use the same 781 listed-position targets from 243 test users. We report accuracy and discussion-specific Macro F1, averaging F1 across the 497 discussion--stance pairs with gold support. For repeated runs, we report the mean and sample standard deviation across seeds. We select checkpoints and intervention strengths using validation Macro F1. Appendix~\ref{app:training} provides method-specific settings and matched controls.

\subsection{Test-Time Scaling Strategies}
\label{sec:main-results}
\label{sec:tts-results}

Table~\ref{tab:tts-main} compares single-sample direct generation with six TTS adaptations. These include Verbalized Sampling, Universal Self-Consistency, AlphaCode-style consensus, PlanSearch, AlphaCodium, and CodeT. Fixed-pool methods use four candidates, whereas AlphaCodium iteratively refines a draft. These comparisons assess prediction strategies without matching total inference compute.

As shown in Table~\ref{tab:tts-main}, additional test-time computation yielded limited Macro F1 gains in these experiments. On Gemini 3.7 Flash, CodeT was the only adaptation to exceed direct generation, reaching 22.63 Macro F1 compared with 22.43. Direct generation achieved the highest Macro F1 on Gemini 3.1 Pro, at 22.05. On Qwen3-8B, Universal Self-Consistency reached 20.66 Macro F1 compared with 20.53 for direct generation. The other Qwen adaptations yielded Macro F1 scores ranging from 16.67 to 20.64. These results complement the candidate-level diagnosis of generation and selection failures in Section~\ref{sec:failure-modes}.

\begin{table}[h]
\centering
\caption{\textbf{Test-time scaling across Gemini and Qwen.} Accuracy (Acc.) and discussion-specific Macro F1 (F1) are percentages on test set. Qwen entries report mean\,$\pm$\,sample s.d.\ over three generation seeds. $k$ counts prediction candidates and AlphaCodium uses iterative refinement. Candidate counts do not equalize total inference compute. Bold marks column maxima.}
\label{tab:main}
\label{tab:tts-main}
\begingroup
\small
\fontfamily{lmr}\selectfont
\renewcommand{\arraystretch}{1.12}
\newcommand{\sd}[1]{\textsubscript{\color{black!55}$\pm$#1}}
\setlength{\tabcolsep}{3.8pt}
\begin{tabular}{@{}lccccccc@{}}
\toprule
& & \multicolumn{2}{c}{Gemini 3.7 Flash} & \multicolumn{2}{c}{Gemini 3.1 Pro} & \multicolumn{2}{c}{Qwen3-8B} \\
\cmidrule(lr){3-4}\cmidrule(lr){5-6}\cmidrule(l){7-8}
Method & $k$ & Acc.$\uparrow$ & F1$\uparrow$ & Acc.$\uparrow$ & F1$\uparrow$ & Acc.$\uparrow$ & F1$\uparrow$ \\
\midrule
Direct generation & 1 & 27.53 & 22.43 & 24.46 & \textbf{22.05} & 26.89\sd{1.30} & 20.53\sd{1.37} \\
\midrule
Verbalized Sampling & 4 & 25.48 & 21.46 & 21.77 & 19.97 & 20.57\sd{0.07} & 16.67\sd{0.47} \\
Universal Self-Consistency & 4 & 27.27 & 22.41 & \textbf{24.58} & 21.69 & \textbf{27.83}\sd{0.27} & \textbf{20.66}\sd{0.24} \\
AlphaCode-style consensus & 4 & 27.53 & 22.30 & \textbf{24.58} & 21.84 & 27.49\sd{0.66} & 20.64\sd{0.42} \\
PlanSearch & 4 & 21.25 & 18.14 & 23.30 & 20.65 & 22.36\sd{1.42} & 18.42\sd{1.49} \\
AlphaCodium & -- & 26.38 & 21.45 & 21.51 & 19.85 & 23.82\sd{1.22} & 18.46\sd{0.84} \\
CodeT & 4 & \textbf{27.78} & \textbf{22.63} & \textbf{24.58} & 21.85 & 26.08\sd{1.28} & 19.98\sd{1.55} \\
\bottomrule
\end{tabular}
\endgroup
\par\vspace{2pt}
\end{table}

\subsection{Learning and Adaptation Strategies}
\label{sec:training-results}

Table~\ref{tab:training-main} compares learning and adaptation strategies using Qwen3-8B as the shared base model. We evaluate LEVER with verifier-only and product selection, SFT, GRPO, user-profile methods, steering vectors, and our method on the test set. The two baselines are direct generation from Table~\ref{tab:tts-main} and direct scoring, which selects the highest-scoring candidate label.

The SFT baseline is trained to generate observed comments and their stance labels. The explicit-profile model includes a textual user profile in its input. The latent-profile model learns a history-conditioned residual edit while keeping the base model frozen. Both profile models use model-generated rationale supervision and additional training tasks constructed from earlier labeled discussions. Steering adds a direction estimated from correct and incorrect training responses to hidden activations of the frozen model. Appendix~\ref{app:training} details the training configurations and their matched controls.

\begin{table}[h]
\centering
\caption{\textbf{Learning and adaptation on Qwen3-8B.} Test-set accuracy and Macro F1 (\%). Subscripted entries report means with sample s.d.\ over three generation seeds (direct generation and steering) or training seeds (SFT, GRPO and profiles). Other entries are deterministic or single fitted systems. Bold marks column maxima, not significance. Settings and matched controls are in Appendix~\ref{app:training}.}
\label{tab:training-main}
\begingroup
\small
\fontfamily{lmr}\selectfont
\renewcommand{\arraystretch}{1.08}
\newcommand{\sd}[1]{\textsubscript{\color{black!65}$\pm$#1}}
\newcommand{\score}[2]{\makebox[2.5em][r]{#1}\makebox[2.7em][l]{#2}}
\setlength{\tabcolsep}{8pt}
\setlength{\aboverulesep}{3pt}
\setlength{\belowrulesep}{3pt}
\arrayrulecolor{black!75}
\begin{tabularx}{\linewidth}{X>{\centering\arraybackslash}p{0.22\linewidth}>{\centering\arraybackslash}p{0.22\linewidth}}
\toprule[0.6pt]
\textbf{Method} & \textbf{Accuracy}$\uparrow$ & \textbf{Macro F1}$\uparrow$ \\
\midrule[0.35pt]
Direct generation & \score{26.89}{\sd{1.30}} & \score{20.53}{\sd{1.37}} \\
Direct scoring & \score{25.86}{} & \score{19.27}{} \\
\addlinespace[3pt]
LEVER (verifier, $k=16$) & \score{27.53}{} & \score{20.13}{} \\
LEVER (product, $k=16$) & \score{25.48}{} & \score{19.78}{} \\
\addlinespace[3pt]
SFT & \score{19.50}{\sd{1.37}} & \score{16.35}{\sd{1.19}} \\
GRPO & \score{27.78}{\sd{0.90}} & \score{21.35}{\sd{0.95}} \\
\addlinespace[3pt]
SFT + user profile & \score{23.60}{\sd{1.15}} & \score{20.74}{\sd{1.49}} \\
Latent profile & \score{23.56}{\sd{3.15}} & \score{21.11}{\sd{2.87}} \\
Steering vector & \score{27.10}{\sd{0.98}} & \score{21.77}{\sd{0.79}} \\
\midrule[0.35pt]
\rowcolor{blue!5}
\textbf{Our method} & \score{\textbf{28.04}}{} & \score{\textbf{21.83}}{} \\
\bottomrule[0.6pt]
\end{tabularx}
\endgroup
\end{table}

The adaptations showed mixed results in discussion-specific Macro F1. SFT reached 16.35, below 21.50 for its matched untrained generator. GRPO reached 21.35 compared with 20.55 for its matched base. LEVER's verifier and product variants scored 20.13 and 19.78, respectively, below 21.14 for majority selection from the same candidate pool. Latent profiles reached 21.11, compared with 20.39 for the matched model without the residual edit.

Profile-inserted SFT reached 20.74 Macro F1 with the target user's profile, compared with 19.23 using another user's profile. Steering reached 21.77 compared with 20.66 for its zero-intervention control. The primary 95\% bootstrap intervals included zero for both the own-minus-donor and steering-minus-control comparisons.

Our method (Section~\ref{sec:method}) achieved the highest accuracy and Macro F1 point estimates in Table~\ref{tab:training-main}, at 28.04\% and 21.83, respectively. Its Macro F1 exceeded direct generation by 1.30 points and direct scoring by 2.56 points. The primary paired comparisons against the direct-branch and different-user evidence controls did not pass Holm correction (Appendix~\ref{app:ours}).

\subsection{Ablation Study}
\label{sec:ablation}

\begin{table}[h]
\centering
\vspace{-3mm}
\caption{\textbf{Ablations of direct and evidence scores on Qwen3-8B.} We report accuracy and Macro F1 (\%) on the test set. Branch ablations set one coefficient to zero. History substitutions keep the direct scores and full-method coefficients fixed. Different-user results average three donor assignments. The final row appends the same retrieved comments to the direct prompt. Shading identifies the full method. Details are in Appendix~\ref{app:ours}.}
\label{tab:method-ablation}
\begingroup
\small
\fontfamily{lmr}\selectfont
\renewcommand{\arraystretch}{1.08}
\newcommand{\sd}[1]{\textsubscript{\color{black!65}$\pm$#1}}
\setlength{\tabcolsep}{8pt}
\setlength{\aboverulesep}{3pt}
\setlength{\belowrulesep}{3pt}
\arrayrulecolor{black!75}
\begin{tabularx}{\linewidth}{X>{\centering\arraybackslash}p{0.18\linewidth}>{\centering\arraybackslash}p{0.18\linewidth}}
\toprule[0.6pt]
\textbf{Variant} & \textbf{Accuracy}$\uparrow$ & \textbf{Macro F1}$\uparrow$ \\
\midrule[0.35pt]
\rowcolor{blue!5}
Full method ($aB+bS$) & 28.04 & 21.83 \\
Direct scores only ($b=0$) & 25.48 & 19.00 \\
Evidence scores only ($a=0$) & 23.94 & 20.06 \\
\midrule[0.35pt]
Different-user history & 25.65 & 19.93 \\
Recent history & 27.91 & 22.30 \\
\addlinespace[3pt]
Retrieved comments in direct prompt & 27.53 & 19.94 \\
\bottomrule[0.6pt]
\end{tabularx}
\endgroup
\vspace{-3mm}
\end{table}

\paragraph{Personal history and retrieval.}
Table~\ref{tab:method-ablation} examines whether the evidence must come from the target user. Both history substitutions keep the direct scores and fused coefficients fixed and change only the comments read by the evidence branch. Replacing the target user's retrieved comments with another user's, averaged over three donor assignments, lowered accuracy from 28.04\% to 25.65\% and Macro F1 from 21.83 to 19.93. This suggests that the evidence branch uses information specific to the target user, although the difference is not statistically conclusive (Appendix~\ref{app:ours}). Selecting the user's most recent comments instead of the most similar ones gave nearly the same accuracy (27.91\%) and slightly higher Macro F1 (22.30), so a simple recency rule is a competitive alternative to similarity-based retrieval.


\paragraph{Method components.}
Table~\ref{tab:method-ablation} also examines how the two score sources are combined. Removing the evidence branch lowered Macro F1 from 21.83 to 19.00, and removing the direct branch lowered it to 20.06. The evidence-only model had lower accuracy than the direct-only model (23.94\% vs.\ 25.48\%) but higher Macro F1, suggesting that the two branches have complementary strengths across stances. Appending the same retrieved comments to the direct prompt kept accuracy close to the full method but reduced Macro F1 to 19.94. Because Macro F1 weights every stance equally, this gap suggests that separate assessment before fusion yields more balanced predictions across stances. Ablation definitions and paired comparisons are provided in Appendix~\ref{app:ours}.

\section{Conclusion}
In this paper, we introduced Stance-Bench to examine the limits of test-time scaling and post-training for individual stance prediction. Our analysis identifies four failure modes: incorrect consensus, selection failure, response overfitting, and early plateau. Guided by these findings, we explored an explicit evidence reader that combines direct stance scores with assessments of personal history, achieving higher Macro F1 than direct scoring while keeping the language model frozen. These exploratory results highlight explicit assessment of personal evidence as a promising direction and motivate evaluating candidate generation, final selection, and person-specific evidence use separately when developing individual social simulations.

\subsection*{AI use statement}
We used generative AI for benchmark annotation, including constructing discussion-specific stance taxonomies and assigning reference labels, as described in Section~\ref{sec:benchmark}. Annotation quality was assessed through the auditing procedures reported in the paper. We also used generative AI tools to assist with code implementation and to improve the language and clarity of the manuscript. The authors take full responsibility for the research and its final presentation, including all AI-assisted annotations, code, text, and claims.

\subsection*{Ethics statement}
This study uses publicly available discussions and comments from Hacker News, collected through its official API with support from the Algolia index. We anonymized user identifiers in the benchmark to protect user privacy. We also recruited volunteer annotators, who participated voluntarily and were fully informed about the purpose of the study and their annotation tasks. The benchmark is intended to evaluate individual stance prediction in a research setting; model predictions should not be treated as verified personal beliefs or used to make consequential decisions about individuals.

\subsection*{Reproducibility statement}
We describe data collection, stance annotation, task construction, and evaluation metrics in Section~\ref{sec:benchmark}. Section~\ref{sec:method} specifies the direct scoring and evidence-fusion procedure, while Section~\ref{sec:overall-performance} describes the experimental comparisons and validation-based model selection. Appendix~\ref{app:prompts} provides the prompt templates, and Appendix~\ref{app:experiments} documents the shared inputs, evaluation protocol, method adaptations, and ablation procedures.



\bibliography{iclr2027_conference}

@article{chen2026synthetic,
  title   = {When Synthetic Users Fail: A Cross-Domain Benchmark of {LLM}-Simulated Human Survey Responses},
  author  = {Chen, Zihan and Zhu, Di and Zheng, Lei Nico},
  journal = {arXiv preprint arXiv:2607.26348},
  year    = {2026},
  url     = {https://arxiv.org/abs/2607.26348},
}

@inproceedings{hu2026simbench,
  title={Simbench: Benchmarking the ability of large language models to simulate human behaviors},
  author={Hu, Tiancheng and Baumann, Joachim and Lupo, Lorenzo and Collier, Nigel and Hovy, Dirk and R{\"o}ttger, Paul},
  booktitle={International Conference on Learning Representations},
  volume={2026},
  pages={28290--28341},
  year={2026}
}

@article{zhang2025verbalized,
  title   = {Verbalized Sampling: How to Mitigate Mode Collapse and Unlock {LLM} Diversity},
  author  = {Zhang, Jiayi and Yu, Simon and Chong, Derek and Sicilia, Anthony
             and Tomz, Michael R. and Manning, Christopher D. and Shi, Weiyan},
  journal = {arXiv preprint arXiv:2510.01171},
  year    = {2025},
  url     = {https://arxiv.org/abs/2510.01171},
}

@inproceedings{lu2026can,
  title={Can LLM Agents Simulate Multi-Turn Human Behavior? Evidence from Real Online Customer Behavior Data},
  author={Lu, Yuxuan and Huang, Jing and Han, Yan and Yao, Bingsheng and Bei, Sisong and Xie, Yaochen and Sang, Yisi and He, Qi and Wang, Dakuo},
  booktitle={Proceedings of the 64th Annual Meeting of the Association for Computational Linguistics (Volume 1: Long Papers)},
  pages={43961--43977},
  year={2026}
}

@article{chen2023universal,
  title={Universal self-consistency for large language model generation},
  author={Chen, Xinyun and Aksitov, Renat and Alon, Uri and Ren, Jie and Xiao, Kefan and Yin, Pengcheng and Prakash, Sushant and Sutton, Charles and Wang, Xuezhi and Zhou, Denny},
  journal={arXiv preprint arXiv:2311.17311},
  year={2023}
}

@article{chen2022codet,
  title={Codet: Code generation with generated tests},
  author={Chen, Bei and Zhang, Fengji and Nguyen, Anh and Zan, Daoguang and Lin, Zeqi and Lou, Jian-Guang and Chen, Weizhu},
  journal={arXiv preprint arXiv:2207.10397},
  year={2022}
}

@article{ridnik2024code,
  title={Code generation with alphacodium: From prompt engineering to flow engineering},
  author={Ridnik, Tal and Kredo, Dedy and Friedman, Itamar},
  journal={arXiv preprint arXiv:2401.08500},
  year={2024}
}

@inproceedings{wang2025planning,
  title={Planning in natural language improves llm search for code generation},
  author={Wang, Evan and Cassano, Federico and Wu, Catherine and Bai, Yunfeng and Song, William and Nath, Vaskar and Han, Ziwen and Hendryx, Sean and Yue, Summer and Zhang, Hugh},
  booktitle={International Conference on Learning Representations},
  volume={2025},
  pages={2432--2478},
  year={2025}
}

@inproceedings{salemi2024lamp,
  title={Lamp: When large language models meet personalization},
  author={Salemi, Alireza and Mysore, Sheshera and Bendersky, Michael and Zamani, Hamed},
  booktitle={Proceedings of the 62nd Annual Meeting of the Association for Computational Linguistics (Volume 1: Long Papers)},
  pages={7370--7392},
  year={2024}
}

@inproceedings{mohammad2016semeval,
  title={Semeval-2016 task 6: Detecting stance in tweets},
  author={Mohammad, Saif and Kiritchenko, Svetlana and Sobhani, Parinaz and Zhu, Xiaodan and Cherry, Colin},
  booktitle={Proceedings of the 10th international workshop on semantic evaluation (SemEval-2016)},
  pages={31--41},
  year={2016}
}

@article{li2022alphacode,
  title   = {Competition-level code generation with {AlphaCode}},
  author  = {Li, Yujia and Choi, David and Chung, Junyoung and Kushman, Nate and Schrittwieser, Julian and Leblond, R{\'e}mi and Eccles, Tom and Keeling, James and Gimeno, Felix and Dal Lago, Agustin and Hubert, Thomas and Choy, Peter and de Masson d'Autume, Cyprien and Babuschkin, Igor and Chen, Xinyun and Huang, Po-Sen and Welbl, Johannes and Gowal, Sven and Cherepanov, Alexey and Molloy, James and Mankowitz, Daniel J. and Sutherland Robson, Esme and Kohli, Pushmeet and de Freitas, Nando and Kavukcuoglu, Koray and Vinyals, Oriol},
  journal = {Science},
  volume  = {378},
  number  = {6624},
  pages   = {1092--1097},
  year    = {2022}
}

@article{shao2024deepseekmath,
  title = {{DeepSeekMath}: Pushing the Limits of Mathematical Reasoning in Open Language Models},
  author = {Shao, Zhihong and Wang, Peiyi and Zhu, Qihao and Xu, Runxin and Song, Junxiao and Bi, Xiao and Zhang, Haowei and Zhang, Mingchuan and Li, Y. K. and Wu, Y. and Guo, Daya},
  journal = {arXiv preprint arXiv:2402.03300},
  year = {2024},
  url = {https://arxiv.org/abs/2402.03300}
}

@inproceedings{ni2023lever,
  title = {{LEVER}: Learning to Verify Language-to-Code Generation with Execution},
  author = {Ni, Ansong and Iyer, Srini and Radev, Dragomir and Stoyanov, Ves and Yih, Wen-tau and Wang, Sida I. and Lin, Xi Victoria},
  booktitle = {Proceedings of the 40th International Conference on Machine Learning},
  pages = {26106--26128},
  year = {2023},
  url = {https://proceedings.mlr.press/v202/ni23b.html}
}

@article{argyle2023out,
  title   = {Out of One, Many: Using Language Models to Simulate Human Samples},
  author  = {Argyle, Lisa P. and Busby, Ethan C. and Fulda, Nancy
             and Gubler, Joshua R. and Rytting, Christopher
             and Wingate, David},
  journal = {Political Analysis},
  volume  = {31},
  number  = {3},
  pages   = {337--351},
  year    = {2023},
  doi     = {10.1017/pan.2023.2},
  url     = {https://doi.org/10.1017/pan.2023.2}
}

@misc{park2024llm,
  title         = {{LLM} Agents Grounded in Self-Reports Enable
                   General-Purpose Simulation of Individuals},
  author        = {Park, Joon Sung and Zou, Carolyn Q.
                   and Kamphorst, Jonne and Egan, Niles
                   and Shaw, Aaron and Hill, Benjamin Mako
                   and Cai, Carrie and Morris, Meredith Ringel
                   and Liang, Percy and Willer, Robb
                   and Bernstein, Michael S.},
  year          = {2026},
  eprint        = {2411.10109},
  archivePrefix = {arXiv},
  primaryClass  = {cs.AI},
  note          = {Version 3; first posted in 2024},
  url           = {https://arxiv.org/abs/2411.10109v3}
}

@inproceedings{santurkar2023whose,
  title     = {Whose Opinions Do Language Models Reflect?},
  author    = {Santurkar, Shibani and Durmus, Esin
               and Ladhak, Faisal and Lee, Cinoo
               and Liang, Percy and Hashimoto, Tatsunori},
  booktitle = {Proceedings of the 40th International Conference
               on Machine Learning},
  series    = {Proceedings of Machine Learning Research},
  volume    = {202},
  pages     = {29971--30004},
  publisher = {PMLR},
  year      = {2023},
  url       = {https://proceedings.mlr.press/v202/santurkar23a.html}
}

@inproceedings{snell2025scaling,
  title     = {Scaling {LLM} Test-Time Compute Optimally Can Be
               More Effective than Scaling Parameters for Reasoning},
  author    = {Snell, Charlie and Lee, Jaehoon
               and Xu, Kelvin and Kumar, Aviral},
  booktitle = {The Thirteenth International Conference
               on Learning Representations},
  year      = {2025},
  url       = {https://openreview.net/forum?id=4FWAwZtd2n}
}

@misc{deepseekai2025deepseekr1,
  title         = {{DeepSeek-R1}: Incentivizing Reasoning Capability
                   in {LLMs} via Reinforcement Learning},
  author        = {{DeepSeek-AI}},
  year          = {2025},
  eprint        = {2501.12948},
  archivePrefix = {arXiv},
  primaryClass  = {cs.CL},
  url           = {https://arxiv.org/abs/2501.12948}
}

@article{gao2024large,
  title={Large language models empowered agent-based modeling and simulation: A survey and perspectives},
  author={Gao, Chen and Lan, Xiaochong and Li, Nian and Yuan, Yuan and Ding, Jingtao and Zhou, Zhilun and Xu, Fengli and Li, Yong},
  journal={Humanities and Social Sciences Communications},
  volume={11},
  number={1},
  pages={1259},
  year={2024},
  publisher={Palgrave}
}

@inproceedings{liu2026cobra,
  author    = {Liu, Xuan and Shang, HaoYang and Jin, Haojian},
  title     = {{CoBRA}: Programming Cognitive Bias in Social Agents Using Classic Social Science Experiments},
  booktitle = {Proceedings of the 2026 CHI Conference on Human Factors in Computing Systems},
  series    = {CHI '26},
  articleno = {64},
  numpages  = {30},
  publisher = {Association for Computing Machinery},
  address   = {New York, NY, USA},
  year      = {2026},
  doi       = {10.1145/3772318.3790804},
  url       = {https://doi.org/10.1145/3772318.3790804}
}

@inproceedings{liu2025exploring,
  title     = {Exploring Prosocial Irrationality for {LLM} Agents: A Social Cognition View},
  author    = {Liu, Xuan and Zhang, Jie and Shang, HaoYang and Guo, Song and Yang, Chengxu and Zhu, Quanyan},
  booktitle = {The Thirteenth International Conference on Learning Representations},
  year      = {2025},
  url       = {https://openreview.net/forum?id=u8VOQVzduP}
}

@inproceedings{zhou2024sotopia,
  title     = {{SOTOPIA}: Interactive Evaluation for Social Intelligence in Language Agents},
  author    = {Zhou, Xuhui and Zhu, Hao and Mathur, Leena and Zhang, Ruohong and Qi, Zhengyang and Yu, Haofei and Morency, Louis-Philippe and Bisk, Yonatan and Fried, Daniel and Neubig, Graham and Sap, Maarten},
  booktitle = {The Twelfth International Conference on Learning Representations},
  year      = {2024},
  url       = {https://openreview.net/forum?id=mM7VurbA4r}
}

@inproceedings{park2023generative,
  title     = {Generative Agents: Interactive Simulacra of Human Behavior},
  author    = {Park, Joon Sung and O'Brien, Joseph C. and Cai, Carrie Jun and Morris, Meredith Ringel and Liang, Percy and Bernstein, Michael S.},
  booktitle = {Proceedings of the 36th Annual ACM Symposium on User Interface Software and Technology},
  series    = {UIST '23},
  articleno = {2},
  numpages  = {22},
  publisher = {Association for Computing Machinery},
  year      = {2023},
  doi       = {10.1145/3586183.3606763},
  url       = {https://doi.org/10.1145/3586183.3606763}
}

@inproceedings{chuang2024beyond,
  title     = {Beyond Demographics: Aligning Role-playing {LLM}-based Agents Using Human Belief Networks},
  author    = {Chuang, Yun-Shiuan and Nirunwiroj, Krirk and Studdiford, Zach and Goyal, Agam and Frigo, Vincent V. and Yang, Sijia and Shah, Dhavan V. and Hu, Junjie and Rogers, Timothy T.},
  booktitle = {Findings of the Association for Computational Linguistics: EMNLP 2024},
  pages     = {14010--14026},
  publisher = {Association for Computational Linguistics},
  year      = {2024},
  doi       = {10.18653/v1/2024.findings-emnlp.819},
  url       = {https://aclanthology.org/2024.findings-emnlp.819/}
}

@inproceedings{anthis2025position,
  title     = {Position: {LLM} Social Simulations Are a Promising Research Method},
  author    = {Anthis, Jacy Reese and Liu, Ryan and Richardson, Sean M. and Kozlowski, Austin C. and Koch, Bernard and Brynjolfsson, Erik and Evans, James and Bernstein, Michael S.},
  booktitle = {Proceedings of the 42nd International Conference on Machine Learning},
  series    = {Proceedings of Machine Learning Research},
  volume    = {267},
  pages     = {81005--81034},
  publisher = {PMLR},
  year      = {2025},
  url       = {https://proceedings.mlr.press/v267/anthis25a.html}
}

@misc{liu2026humanstudy,
  title         = {{HumanStudy-Bench}: Towards {AI} Agent Design for Participant Simulation},
  author        = {Liu, Xuan and Shang, Haoyang and Liu, Zizhang and Liu, Xinyan and Xiao, Yunze and Tu, Yiwen and Jin, Haojian},
  year          = {2026},
  eprint        = {2602.00685},
  archivePrefix = {arXiv},
  primaryClass  = {cs.AI},
  url           = {https://arxiv.org/abs/2602.00685}
}

@inproceedings{dominguez2024questioning,
  title     = {Questioning the Survey Responses of Large Language Models},
  author    = {Dominguez-Olmedo, Ricardo and Hardt, Moritz and Mendler-D{\"u}nner, Celestine},
  booktitle = {Advances in Neural Information Processing Systems},
  volume    = {37},
  year      = {2024},
  url       = {https://proceedings.neurips.cc/paper_files/paper/2024/hash/515c62809e0a29729d7eec26e2916fc0-Abstract-Conference.html}
}

@article{bisbee2024synthetic,
  title     = {Synthetic Replacements for Human Survey Data? The Perils of Large Language Models},
  author    = {Bisbee, James and Clinton, Joshua D. and Dorff, Cassy and Kenkel, Brenton and Larson, Jennifer M.},
  journal   = {Political Analysis},
  volume    = {32},
  number    = {4},
  pages     = {401--416},
  year      = {2024},
  doi       = {10.1017/pan.2024.5},
  url       = {https://doi.org/10.1017/pan.2024.5}
}

@techreport{openai2026gpt55,
  author      = {OpenAI},
  title       = {GPT-5.5 System Card},
  institution = {OpenAI},
  year        = {2026},
  month       = apr,
  url         = {https://deploymentsafety.openai.com/gpt-5-5/gpt-5-5.pdf},
  note        = {Accessed: 2026-09-23}
}

@article{genest1984aggregating,
  title={Aggregating opinions through logarithmic pooling},
  author={Genest, Christian and Weerahandi, Samaradasa and Zidek, James V},
  journal={Theory and decision},
  volume={17},
  number={1},
  pages={61},
  year={1984},
  publisher={Kluwer Academic Publishers}
}

@incollection{McFadden1974,
  author    = {McFadden, Daniel},
  title     = {Conditional Logit Analysis of Qualitative Choice Behavior},
  booktitle = {Frontiers in Econometrics},
  editor    = {Zarembka, Paul},
  pages     = {105--142},
  publisher = {Academic Press},
  address   = {New York},
  year      = {1974}
}

@inproceedings{wang2023selfconsistency,
  title     = {Self-Consistency Improves Chain of Thought Reasoning in Language Models},
  author    = {Wang, Xuezhi and Wei, Jason and Schuurmans, Dale and Le, Quoc V. and Chi, Ed H. and Narang, Sharan and Chowdhery, Aakanksha and Zhou, Denny},
  booktitle = {The Eleventh International Conference on Learning Representations},
  year      = {2023},
  url       = {https://openreview.net/forum?id=1PL1NIMMrw}
}

@article{loh2024predicting,
  title={Predicting user stances from target-agnostic information using large language models},
  author={Loh, Siyuan Brandon and Wong, Liang Ze and Bhattacharya, Prasanta and Simons, Joseph and Gao, Wei and Zhang, Hong},
  journal={arXiv preprint arXiv:2409.14395},
  year={2024}
}

@article{ouyang2022training,
  title={Training language models to follow instructions with human feedback},
  author={Ouyang, Long and Wu, Jeffrey and Jiang, Xu and Almeida, Diogo and Wainwright, Carroll and Mishkin, Pamela and Zhang, Chong and Agarwal, Sandhini and Slama, Katarina and Ray, Alex and others},
  journal={Advances in neural information processing systems},
  volume={35},
  pages={27730--27744},
  year={2022}
}
\bibliographystyle{iclr2027_conference}

\clearpage
\appendix
\section{Diagnostic Coverage of Existing Evaluations}
\label{app:diagnostic-coverage}

Existing evaluations address complementary aspects of person-specific prediction. Table~\ref{tab:comparison} compares their reported protocols using the three diagnostic requirements introduced in Section~\ref{sec:motivation-diagnosis}, alongside person-level targets and naturally occurring observations. The entries describe the evaluation settings as reported; a question mark indicates that the relevant support is unclear.

\begin{table}[htbp]
\centering
\footnotesize
\setlength{\tabcolsep}{2pt}
\renewcommand{\arraystretch}{1.0}
\caption{\small Diagnostic coverage of existing evaluations and \benchmark.
R1: users compared within a shared context;
R2: personal evidence removed or replaced with the prediction target fixed;
R3: candidate and final predictions recorded separately in a shared stance space.
ULSP (CB): user-level stance prediction on Connected Behaviour.
\cmark: supported; \pmark: partial; \xmark: unsupported;
?: unclear from the reported evaluation.}
\label{tab:comparison}
\begin{tabular*}{\linewidth}{@{\extracolsep{\fill}}lccccc@{}}
\toprule
Evaluation
& \shortstack{Person-level\\targets}
& \shortstack{Shared\\context (R1)}
& \shortstack{Evidence\\controls (R2)}
& \shortstack{Candidate/final\\traces (R3)}
& \shortstack{Naturally\\occurring} \\
\midrule
OpinionQA \citep{santurkar2023whose}
    & \xmark & \cmark & \xmark & \xmark & \xmark \\
SimBench \citep{hu2026simbench}
    & \xmark & \cmark & \xmark & \xmark & \pmark \\
\citet{park2024llm}
    & \cmark & \cmark & ? & \xmark & \xmark \\
LaMP \citep{salemi2024lamp}
    & \cmark & \xmark & ? & \xmark & \cmark \\
SemEval-2016 \citep{mohammad2016semeval}
    & \xmark & \pmark & \xmark & \xmark & \cmark \\
ULSP (CB) \citep{loh2024predicting}
    & \cmark & \pmark & ? & \xmark & \cmark \\
\midrule
\textbf{\benchmark (ours)}
    & \cmark & \cmark & \cmark & \cmark & \cmark \\
\bottomrule
\end{tabular*}
\end{table}

\benchmark combines these requirements within a common evaluation setting. Multiple users are linked to the same naturally occurring discussion, and their preceding histories can be removed or replaced while the discussion and reference stance remain fixed. For methods that generate multiple candidates, the evaluation records each candidate's stance label and the final decision separately. Together, these properties support controlled analyses of personal evidence and distinguish candidate coverage from final selection.

\section{Prompt Templates}
\label{app:prompts}

The box below presents the prompt templates used for stance prediction, historical evidence reading, and label-guided comment generation. Headings identify stages and message roles.

\begin{promptbox}{PromptBlue}{Prompt templates}
[Stance prediction: System prompt]
You are simulating one specific Hacker News user (the subject) from their comment history. Predict how they will respond in a new discussion.
Return JSON with:
- reasoning: at most three sentences connecting their history to this prediction.
- predicted_comment: the comment this subject would most plausibly write, in their own voice and typical length. Write the comment itself, never a description of it.
- label_id: the candidate position that comment takes. Use OTHER if their position is not listed.
The comment and the label must express the same stance.

[Stance prediction: User prompt]
== SUBJECT HISTORY: their 8 most recent labeled discussions, oldest first ==
[1] Discussion: <historical_title>  (<days_before> days before)
    Argued over: <historical_central_question>
    Position the subject took: <position_taken>
    Positions they did not take: <alternative_1> | <alternative_2> | ...
    Subject wrote: "<historical_comment>"
<remaining_history_entries_in_the_same_format>

== NEW DISCUSSION the subject is about to comment in ==
Title: <target_title>
Link domain: <link_domain>
Posted: <YYYY-MM-DD>
Story text:
<story_text>
Comments already in the thread (as a visitor would first see it):
  - <visible_comment_1>
  - <visible_comment_2>
  ...
The discussion's central question: <target_central_question>

== CANDIDATE POSITIONS: the subject's comment will take exactly one (listed in no particular order) ==
<candidate_rows>

[Direct label scoring: Assistant prefix]
{"label_id": "

[Direct label scoring: Candidate continuation]
<candidate_id>", "

[Historical evidence reading: System prompt]
You infer a Hacker News user's likely position in a discussion from one comment they wrote elsewhere. Answer in JSON.

[Historical evidence reading: User prompt]
Central question: <target_central_question>
Discussion: <target_title>

Positions:
[A] <position_name_A>: <definition_A> (Not this if: <boundary_A>)
[B] <position_name_B>: <definition_B> (Not this if: <boundary_B>)
...
[U] The comment says nothing about which of these positions this person holds.

This person wrote the following comment earlier, elsewhere on Hacker News:
"<retrieved_historical_comment>"

Which of the positions above is this person most likely to take in the discussion? Return JSON {"answer": "<letter>"}.

[Historical evidence reading: Assistant prefix]
{"answer": "

[Label-guided comment generation: Append to prediction user prompt]
The subject's position in this discussion has been determined to be [<predicted_label_id>] <predicted_label_name>. Write the comment they would post taking that position.

For this step, write only the subject's comment itself as plain text: no JSON, no reasoning, no label, no quotation marks around it.
\end{promptbox}
\section{Method Adaptations and Experimental Details}
\label{app:experiments}

\subsection{Shared inputs and evaluation}
\label{app:inputs}

Methods receive the target discussion, its candidate stance definitions, and the user's preceding annotated discussions. The main comparisons use the eight most recent eligible historical discussions. Generative methods produce a predicted comment and a declared stance label; direct-scoring methods select a label without generating a comment. User histories precede the user's first contribution to the target discussion. Comments by any cohort user in the target discussion, including the target user, are excluded from the model-visible target context.

\paragraph{Evaluation.}
Accuracy is the fraction of the 781 test targets whose predicted label matches the reference stance. Discussion-specific Macro F1 treats each discussion--stance pair with at least one reference target as a class. For each pair, we count true positives, false positives, and false negatives over targets in that discussion and compute $F_1 = 2\mathrm{TP}/(2\mathrm{TP}+\mathrm{FP}+\mathrm{FN})$. We average $F_1$ over the 497 such pairs. Invalid outputs count as false negatives for the reference stance and do not add false positives.

\subsection{Direct prediction and test-time scaling}
\label{app:methods}

The TTS adaptations use the corresponding frozen backbone in Table~\ref{tab:tts-main}; Direct scoring is evaluated on Qwen3-8B. Fixed-pool methods use four responses, while AlphaCodium refines a single draft. Predictions are returned in structured form with a comment and its declared stance label. For methods originating in code generation, we replace programs with predicted comments and implement selection or checking through stance agreement and response properties.

\paragraph{Direct generation and Direct scoring.}
Direct generation makes one call to produce a comment in the user's voice, a short rationale grounded in the history, and a stance label consistent with the comment. Its declared label is the final prediction. Direct scoring instead appends each candidate label continuation to a shared prediction prompt with a fixed label-first output prefix. It sums the conditional token log-probabilities of each continuation, including its closing delimiter, and selects the highest-scoring label under the frozen model.

\paragraph{Verbalized Sampling.}
We ask the model to propose four diverse plausible responses and assign a probability to each in a single call \citep{zhang2025verbalized}. Each response contains a comment, its stance label, and a brief explanation connecting the prediction to the user's history. The prompt encourages distinct predictions and asks for probabilities that approximately sum to one. We normalize the returned probabilities over the four responses and select the response with the highest value, resolving ties by candidate order.

\paragraph{Universal Self-Consistency.}
We independently sample four comment-and-label predictions and make an additional selection call to the same backbone \citep{chen2023universal}. The selector receives the original input together with the indexed candidate labels and comments, with each comment limited to 700 characters. It identifies the candidate whose stance and supporting reasoning best reflect agreement across the responses and returns its index with a short explanation. Selection uses temperature zero, and the chosen candidate's declared label is the final prediction. The procedure uses five model calls.

\paragraph{AlphaCode-style consensus.}
We adapt behavioral clustering \citep{li2022alphacode} by independently generating four responses and grouping them by their declared stance labels. Group size measures agreement on the predicted position. We select the first candidate in the largest group; when groups have equal size, the group containing the earliest candidate wins. This procedure uses four generation calls, with grouping and selection performed directly from their labels.

\paragraph{PlanSearch.}
We adapt the observation--hypothesis--generation pipeline to comment prediction \citep{wang2025planning}. The model first extracts four to eight observations from the user's history, covering values, recurring concerns, and argumentative habits. Using these observations and the target discussion, it then proposes four distinct hypotheses, each comprising a stance summary and a short argument sketch. Each hypothesis conditions a separate call that produces a comment, rationale, and label; the label is assigned according to the generated comment. The four responses are selected by the same label-consensus rule as above, including its ordering-based tie rule. Observation extraction uses temperature 0.7, and hypothesis and response generation use temperature 0.95. The procedure uses six model calls.

\paragraph{AlphaCodium.}
We adapt reflection and iterative repair to predicted comments \citep{ridnik2024code}. The model first summarizes the user's characteristic views and argumentative habits, interprets the discussion, and ranks up to four plausible positions. This reflection conditions an initial comment and stance prediction. A separate call derives three to six acceptance checks from the history and discussion, covering stance, argument type, register, typical length, and writing habits. The model then performs up to two rounds of checking and revision: each round identifies failed checks and returns a revised comment and label while preserving properties already satisfied. The final draft is returned. Reflection uses temperature 0.4, check generation uses 0.6, and initial prediction and revision use 0.95. The procedure uses three preparatory calls and up to two repair calls.

\paragraph{CodeT.}
We independently generate four candidate responses and a separate set of yes/no assertions describing properties of the comment the user is expected to write \citep{chen2022codet}. Assertion generation receives the original input without the candidate responses. Each candidate is then checked in a separate call containing its comment and the numbered assertions; the checker returns the indices of satisfied assertions at temperature zero. Candidates satisfying the same assertion set form a group $G$, scored by $|G|\,|A_G|$, where $A_G$ is their shared set of satisfied assertions. We return the first candidate in the highest-scoring group, breaking score ties by group size and then candidate order. We use eight, four, and three assertions for Qwen3-8B, Gemini Flash, and Gemini Pro, respectively. The procedure uses nine calls: four for candidates, one for assertions, and four for checking.

\paragraph{Decoding settings.}
Direct generation uses temperature 0.2; TTS candidate generation and CodeT assertion generation use temperature 0.95. Stage-specific temperatures are given above. For Qwen3-8B, the output cap per call is 3,072 tokens for Verbalized Sampling, 1,536 for AlphaCodium, and 1,024 for Direct generation and the remaining TTS methods. Gemini Pro uses a 32,000-token output cap per call. Qwen3-8B runs use the non-thinking chat template and generation seeds 17, 18, and 19.

\subsection{Learning and adaptation strategies}
\label{app:training}

The learning and adaptation baselines share the Qwen3-8B backbone and use an eight-discussion history window. SFT, GRPO, and profile-method checkpoints, together with steering strengths, are selected using validation Macro F1.

All trained LoRA adapters target the attention query, key, value, and output projections and the feed-forward gate, up, and down projections. SFT, SFT + user profile, and LEVER use rank $r=64$, scaling parameter $\alpha=128$, and dropout 0.05. GRPO uses $r=32$, $\alpha=64$, and zero dropout. Latent profile keeps the backbone frozen, and Steering vector requires no gradient-based training. Table~\ref{tab:baseline-decoding} specifies test decoding; maximum generation lengths count newly generated tokens, whereas training sequence limits include both prompt and response.

\paragraph{LEVER.}
We adapt learned verification \citep{ni2023lever} by training a verifier with LoRA adapters and a classification head to predict whether a response's declared label matches the observed stance. It receives the task context and candidate response; label correctness replaces program-execution feedback. Both variants rank the same pool of 16 responses. LEVER (verifier) selects the highest predicted correctness score; LEVER (product) multiplies that score by the exponentiated, length-normalized generation log-probability before selection.
We train for one epoch with AdamW at learning rate $10^{-4}$, zero weight decay, 5\% linear warmup followed by linear decay, and a maximum sequence length of 4,608 tokens. Two devices each process two pairs per microbatch, with eight accumulation steps, giving an effective batch of 32 pairs. The loss is binary cross-entropy with positive-class weight 3.55.

\paragraph{SFT.}
We train LoRA adapters to generate the observed user comment followed by its stance label, conditioned on the discussion and preceding history. At test time, we generate a response and evaluate its declared label, preserving the comment-and-label formulation studied in Section~\ref{sec:sft-failure}.
Training runs for three epochs at learning rate $10^{-4}$, with cosine decay and 5\% warmup. The per-device batch is one task, accumulated over eight steps on one device (effective batch eight); the maximum sequence length is 5,120 tokens. Loss is computed on response tokens. We evaluate each epoch-end checkpoint using generated stance predictions on validation data.

\paragraph{GRPO.}
We adapt group-relative reward-based learning \citep{shao2024deepseekmath} to stance attribution by sampling multiple responses to each training prompt and rewarding exact agreement with the observed stance label. Rewards are $+1$ for a correct label, $0$ for a valid incorrect label, and $-1$ for an invalid output. LoRA policy updates use within-prompt reward differences with KL coefficient $\beta=0.02$ toward the frozen base model. We train for 300 updates using AdamW at constant learning rate $3\times10^{-5}$, optimizer coefficients $(0.9,0.99)$, zero weight decay, and gradient clipping at 1.0. Each update accumulates eight prompts with eight responses per prompt (64 responses). Rollouts use temperature 1.0, top-$p=1.0$, no top-$k$ truncation, and at most 384 new tokens. Checkpoints are evaluated every 50 updates, sampling four responses per validation prompt with the same decoding settings and scoring only the first. Test generation samples eight responses with the settings in Table~\ref{tab:baseline-decoding}, again scoring only the first response.

\paragraph{SFT + user profile.}
We summarize each user's preceding history as a textual profile of recurring preferences and behavioral tendencies, supported by historical evidence, and insert it into the predictor's prompt. LoRA training uses generated rationales together with observed comments and stance labels. Rationales come from label-correct samples or explanations generated with the observed outcome supplied during training.
Profiles are generated by Qwen3-8B at temperature zero from the eight historical discussions alone; the inserted profile retains up to two supporting quotations per field. We train with AdamW at learning rate $10^{-4}$ for two epochs, with 3\% warmup, zero weight decay, and gradient clipping at 1.0. The per-device batch is one task with 32 accumulation steps on one device (effective batch 32); the sequence limit is 6,144 tokens. Checkpoints are evaluated every half epoch.

\paragraph{Latent profile.}
We encode historical comments with the frozen backbone, pool their representations into a user vector, and train a lightweight controller to convert that vector into a residual hidden-state edit. The backbone remains frozen. The profile module is trained with rationale, comment, and label supervision together with an auxiliary training-user identification objective. At test time, the edit is computed from the unseen user's history.
The controller uses a 512-dimensional user representation and a rank-eight edit at decoder layer 8 (zero-indexed), with each historical comment capped at 256 tokens. We train only the profile module and auxiliary head for three epochs at learning rate $10^{-3}$, using a one-task microbatch, 32 accumulation steps, and a 6,144-token sequence limit. The objective adds user-identification cross-entropy with weight 0.5 and an orthogonality penalty with weight $10^{-3}$ to response-token loss. Prediction checkpoints are evaluated every half epoch. SFT and both profile methods use no additional KL penalty.

\paragraph{Steering vector.}
We estimate a shared activation direction by contrasting correct and incorrect stance responses generated for the same training tasks. At inference, we add a validation-selected multiple of this direction to the frozen model's hidden activations. This is a shared correct-position direction, whereas Latent profile produces an edit conditioned on each user's history.
The direction is estimated at decoder layer 20 (zero-indexed). We normalize the mean task-paired activation difference to unit length and test multipliers $f\in\{0,0.15,0.4,0.7\}$, scaled by the training activation norm 88.89. Validation selects $f=0.15$, corresponding to the recorded additive coefficient 13.33. The intervention is applied at the final prompt token and subsequent generated tokens.

\begin{table}[htbp]
\centering
\small
\setlength{\tabcolsep}{4pt}
\renewcommand{\arraystretch}{1.12}
\caption{\textbf{Test decoding for Qwen3-8B baselines.} $T$ denotes temperature; $p$ and $k_{\mathrm{top}}$ denote nucleus and top-$k$ sampling. Token caps apply per response. ``Default'' denotes the generation setting inherited from the loaded model or server configuration.}
\label{tab:baseline-decoding}
\begin{tabularx}{\linewidth}{@{}lccccX@{}}
\toprule
Method & $T$ & $p$ & $k_{\mathrm{top}}$ & Token cap & Prediction readout \\
\midrule
Direct generation & 0.2 & Default & Default & 1,024 & One response \\
LEVER (both variants) & 0.95 & Default & Default & 512 & Rank 16 responses \\
SFT / SFT + user profile & 0.95 & 0.95 & 20 & 2,048 & One response \\
GRPO & 1.0 & 0.95 & 20 & 384 & First of eight responses \\
Latent profile & 0.95 & 1.0 & Default & 2,048 & One response \\
Steering vector & 0.95 & 1.0 & Default & 2,048 & One response \\
\bottomrule
\end{tabularx}
\end{table}

Qwen3-8B generation uses the non-thinking chat template. Direct generation and LEVER candidate sampling use schema-constrained JSON through vLLM; SFT, SFT + user profile, and GRPO use unconstrained vLLM generation, while Latent profile and Steering vector use Hugging Face generation. Direct scoring deterministically selects the label with the highest conditional likelihood. SFT, GRPO, and both profile methods are trained with seeds 1, 2, and 3. Their selected checkpoints are, respectively, epochs 1/3/1, updates 250/300/200, epoch 1 for all three profile-SFT runs, and half an epoch for all three latent-profile runs. Direct generation and Steering vector use generation seeds 17, 18, and 19; the trained vLLM models use test-generation seed 17, and latent-profile generation uses $17+$ batch index.

\subsection{Our method and its ablations}
\label{app:method}
\label{app:ours}

\paragraph{Retrieving and reading personal evidence.}
The direct branch scores labels using the discussion and annotated history. The evidence branch separately retrieves up to eight comments from the user's public history by similarity to the target discussion, retaining at most two per historical discussion \citep{salemi2024lamp}. Retrieved comments precede the user's first target-discussion contribution, exclude the target discussion, and require no historical stance annotations. The frozen reader assesses each comment against the target stance definitions and an additional option $U$ for insufficient evidence, giving a distribution $q_i$ over $\mathcal{L}_d\cup\{U\}$. We average aligned log-probabilities of $q_i$ across option orders and renormalize over listed stances to obtain $p_i$. The weight is $w_i=1-\exp\{\overline{\log q_i(U)}\}$, where the bar denotes the same order averaging. These quantities define the weighted log-probability mean in Eq.~\ref{eq:evidence-score}.

\paragraph{Fusion and prediction.}
Only the two fusion coefficients are fitted, using regularized stance negative log-likelihood on development data; both language models remain frozen. Evidence scores are centered before fusion, and the fitted coefficients are fixed during inference. If no usable evidence is available, the full method uses zero evidence scores.

\paragraph{Matched ablations.}
Direct scores only and Evidence scores only remove the evidence and direct branches, respectively. Their remaining coefficients are fitted on development data; using the corresponding full-method coefficients gives identical predictions because these fitted coefficients are positive. Evidence-only prediction never falls back to the direct branch. Different-user history replaces only the evidence source, while Recent history replaces relevance retrieval with the latest eligible comments; both retain the full-method coefficients and direct scores. Retrieved comments in the direct prompt appends the same comments to the direct predictor instead of scoring them separately. The Direct scoring baseline and the direct-only ablation use different label-scoring implementations, so each is retained as its own reported comparison.

The primary paired Macro F1 differences are $+2.83$ points against Direct scores only (95\% interval $[0.14,5.28]$) and $+1.89$ against Different-user history ($[-0.17,3.90]$). Neither passes Holm correction across these two comparisons. The differences against Evidence scores only ($[-0.36,4.69]$) and Retrieved comments in direct prompt ($[-0.97,4.07]$) also have intervals including zero.

\end{document}